\documentclass[runningheads]{llncs}
\usepackage[utf8x]{inputenc}
\usepackage{graphicx}
\usepackage[font=footnotesize]{caption}
\usepackage[caption=false,font=footnotesize]{subfig}
\usepackage{mdwmath}
\usepackage{mdwtab}
\usepackage{amsmath,amssymb} 
\usepackage{color}
\usepackage[width=122mm,left=12mm,paperwidth=146mm,height=193mm,top=12mm,paperheight=217mm]{geometry}
\graphicspath{{Images/}} 
\usepackage{algorithm}
\usepackage{algpseudocode}
\usepackage{array,multirow}
\usepackage{hyperref}
\usepackage{wrapfig}
\usepackage{hyperref}

\begin{document}
\pagestyle{headings}
\mainmatter
\def\ECCV18SubNumber{1134}  

\title{Localization Recall Precision (LRP): A New Performance Metric for Object Detection} 

\titlerunning{LRP: A New Performance Metric for Object Detection}

\authorrunning{Kemal Oksuz, Baris Can Cam, Emre Akbas and Sinan Kalkan}

\author{Kemal Oksuz, Baris Can Cam, Emre Akbas and Sinan Kalkan}
\institute{Department of Computer Engineering, Middle East Technical University \\ \texttt{\{kemal.oksuz,can.cam,eakbas,skalkan\}@metu.edu.tr}}
\maketitle 
\begin{abstract}

Average precision (AP), the area under the recall-precision (RP) curve, is the standard performance measure for object detection. Despite its wide acceptance, it has a number of shortcomings, the most important of which are (i) the inability to distinguish very different RP curves, and (ii) the lack of directly measuring bounding box localization accuracy. In this paper, we propose ``Localization Recall Precision (LRP) Error’’, a new metric which we specifically designed for object detection. LRP Error is composed of three components related to localization, false negative (FN) rate and false positive (FP) rate. Based on LRP, we introduce the ``Optimal LRP’’, the minimum achievable LRP error representing the best achievable configuration of the detector in terms of recall-precision and the tightness of the boxes. In contrast to AP, which considers precisions over the entire recall domain, Optimal LRP determines the ``best'' confidence score threshold for a class, which balances the trade-off between localization and recall-precision. In our experiments, we show that, for state-of-the-art (SOTA) object detectors, Optimal LRP provides richer and more discriminative information than AP. We also demonstrate that the best confidence score thresholds vary significantly among classes and detectors. Moreover, we present LRP results of a simple online video object detector which uses a SOTA still image object detector and show that the class-specific optimized thresholds increase the accuracy against the common approach of using a general threshold for all classes. At \url{https://github.com/cancam/LRP} we provide the source code that can compute LRP for the PASCAL VOC and MSCOCO datasets. Our source code can easily be adapted to other datasets as well.

\end{abstract}

\section{Introduction}
\label{sec:intro}
Today ``average precision’’ (AP) is the de facto standard for performance evaluation in object detection. The official success criteria in all major object detection datasets and competitions \cite{COCO,ILSVRC,PASCAL} are based on AP. Popular still-image object detection \cite{SSD,FasterRCNN,RFCN,FocalLoss}, video object detection \cite{TCNN,DetectToTrack,MicrosoftVideo} and online video object detection \cite{Context,AssociationLSTM} papers mainly report AP and mean-AP (mAP; explained in Section \ref{sec:AveragePrecision}) results. AP not only enjoys such vast acceptance but it also appears to be unchallenged. Except for a small number of papers which do ablation studies \cite{FasterRCNN,FocalLoss}, AP appears to be the sole criterion used to compare object detection methods.

\begin{figure}[hbt!]
\centering
\begin{tabular}{ccc}
\includegraphics[width=0.31\textwidth]{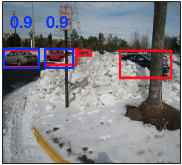} & 
\includegraphics[width=0.31\textwidth]{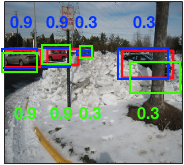} &  
\includegraphics[width=0.31\textwidth]{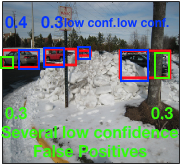} \\
\scriptsize (a) & \scriptsize (b) &  \scriptsize (c) \\
\includegraphics[width=0.31\textwidth]{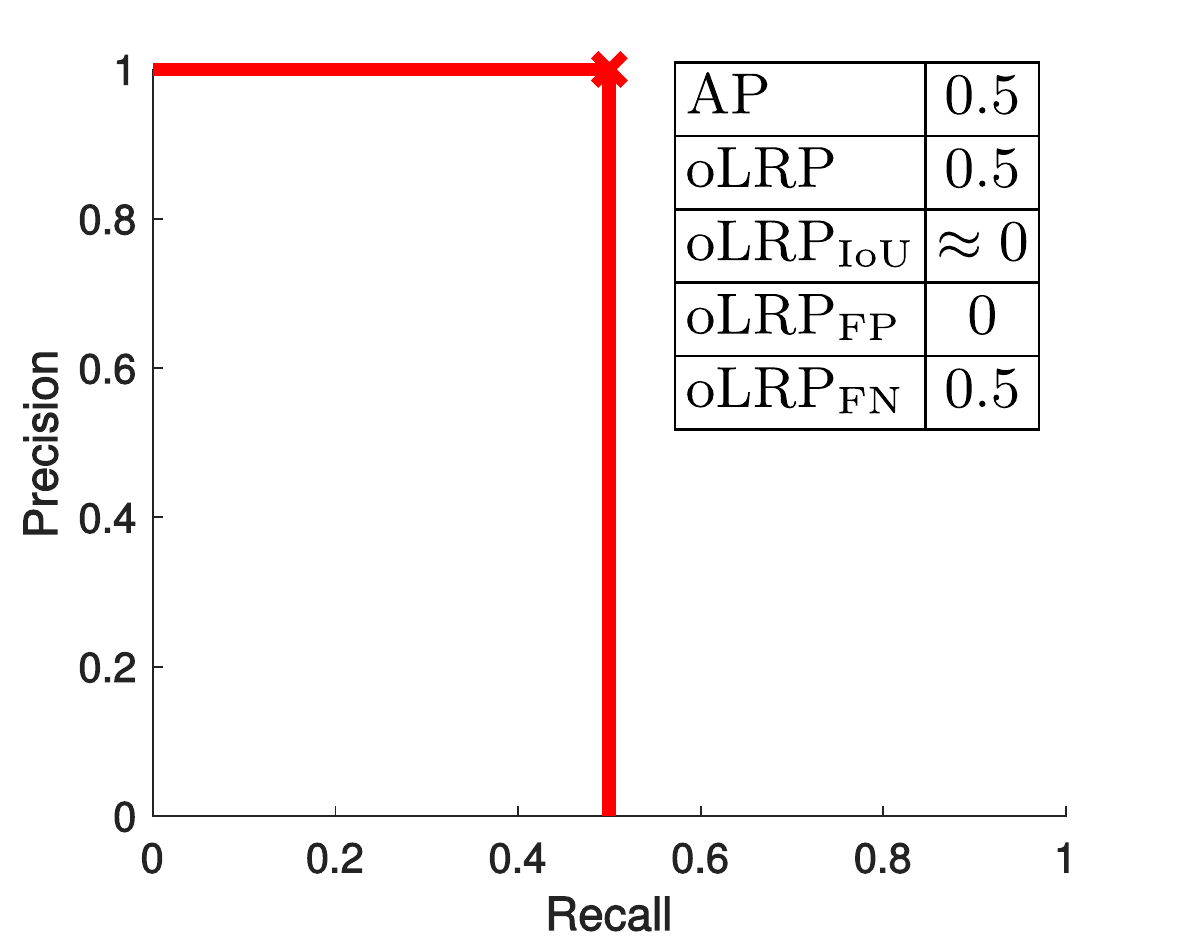} & 
\includegraphics[width=0.31\textwidth]{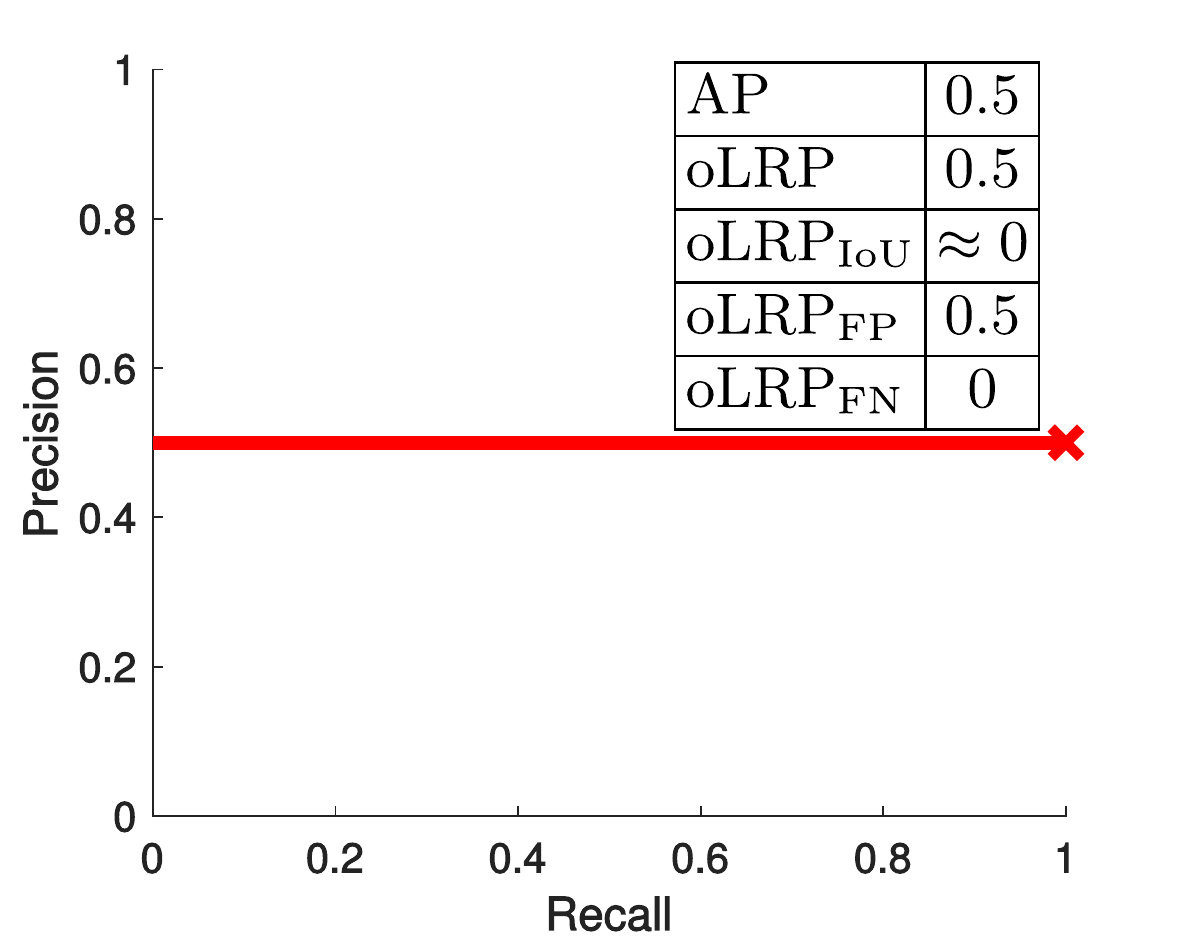} & 
\includegraphics[width=0.31\textwidth]{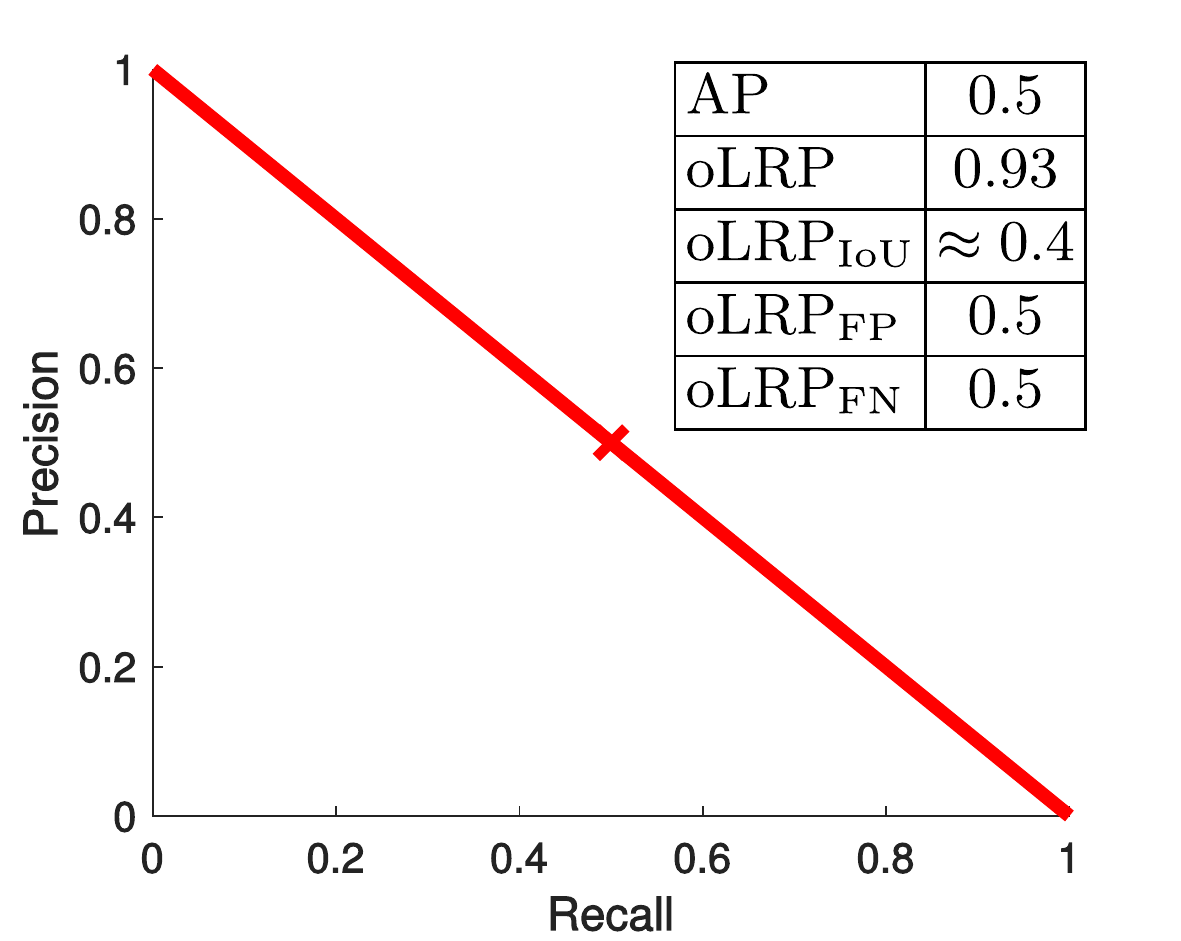} \\
\scriptsize (d) & \scriptsize (e) &  \scriptsize (f) \\
\end{tabular}
\caption{Three different object detection results (for an image from ILSVRC \cite{ILSVRC}) with very different RP curves but the same AP. AP is unable to identify the difference between these curves. \textbf{(a,b,c)} Red, blue and green colors denote ground-truth, true positives; false positives respectively. Numbers are detection confidence scores. \textbf{(d,e,f)} RP curves, AP and LRP results for the corresponding detections in (a,b,c). Red crosses indicate Optimal LRP points. [Best viewed in color]
}
\label{fig:differentdetectors}
\end{figure}

Despite its popularity, AP has certain deficiencies. First, \textbf{AP cannot distinguish between very different RP curves}:  In Fig. \ref{fig:differentdetectors}, we present detection results of three hypothetical object detectors, on the same image. The detector in (a) detects only half of the objects but with full precision; this is a low-recall-high-precision detector. In contrast,  the detector in (b) detects all objects; however, for each correct detection it also produces a close-to-duplicate detection which escapes non-maxima suppression. Hence, detector (b) is a high-recall-low-precision detector. And the detector in (c) appears to be in between; it represents a detector with higher precision at lower recall and vice versa. Despite these very different characteristics, the APs of these differently-behaving detectors are exactly the same (AP=$0.5$). One needs to inspect the RP curves in order to understand the differences in behavior, which can be time-consuming and impractical with large number of classes such as in the ImageNet object detection challenge \cite{ILSVRC} with $200$ classes.

Another deficiency of AP is that \textbf{it does not explicitly include localization accuracy}: One cannot infer from AP the tightness level of the bounding box detections. Nevertheless, since extracting tighter bounding boxes is a desired property, nearly every paper on the topic discusses the issue mostly qualitatively \cite{SSD,FasterRCNN,RFCN,DetectToTrack,AssociationLSTM} and some quantitatively by computing AP scores for different intersection-over-union (IoU) thresholds \cite{SSD,FocalLoss,FasterRCNN}. However, this quantitative approach does not directly measure the localization accuracy either (but it gives an idea about it) and for the qualitative approach, it is very likely for the sample boxes to be very limited and biased. We discuss other less severe deficiencies of AP in Section \ref{sec:AveragePrecision}.


A desirable performance measure (or metric) is expected to include all of the factors related with performance. In object detection, the most important three factors are (i) the localization accuracy of the true positives (TP), (ii) the false positives (FP) rate and (iii) the false negative (FN) rate. Being able to represent the strengths and weaknesses of a detector, based on these factors,  is another desirable property for a performance measure since it can reveal improvement directions. Furthermore, a performance metric should reveal the RP characteristics of a detector (as LRP achieves in Fig. \ref{fig:differentdetectors}). This ability would benefit certain applications. For instance, using a high-precision detector is common in visual tracking methods \cite{Threshold1,Threshold2,Threshold3,Threshold4,Threshold5}, while initializing the tracker, known as \textit{tracking by detection} as faster response times are required. Also, in online video object detection, the current approach is to use a still-image object detector with a general threshold (e.g., Association-LSTM \cite{AssociationLSTM} uses SSD \cite{SSD} detections with confidence score above $0.8$).  A desirable performance measure should help in setting an optimal confidence score threshold per class.


In this paper, we propose a new metric called the ``Localization-Recall-Precision Error’’ (LRP, for short). LRP  involves appropriate components closely related to the precision, recall and IoU overlap and each parametrization of LRP corresponds to a point on the RP curve. We propose the ``Optimal LRP’’, the minimum achievable LRP error, as the alternative performance metric to AP. Optimal LRP alleviates the drawbacks of AP, represents the tightness of the bounding-boxes and the shape of the RP curve via its components and is more suitable for ablation studies. Finally, based on Optimal LRP, a confidence score thresholding method is proposed to decrease the number of detections in an optimal manner.



We conducted three sets of experiments in order to backup our claims. The first analyzes the parameters of LRP. In the second one, we computed both AP and Optimal LRP for common, state-of-the-art object detectors \cite{SSD,FasterRCNN,FocalLoss} on MSCOCO \cite{COCO}. We showed that Optimal LRP provides richer and more discriminative information than AP. In the third set, we built a simple, online video object detector that uses a still-image object detector (RetinaNet \cite{FocalLoss}) as its backbone. We showed that the class-specific confidence score thresholds which are provided by Optimal LRP improved the  performance against the common approach of using a general threshold for all classes.


\section{Related Work}
\label{sec:Related}

Here we review the performance measures related to recall-precision and/or object detection. Since AP is a central topic in this paper, we did not limit our discussion on AP here in a subsection but instead, we discuss several aspects of AP throughout the paper.

\textbf{Information Theoretic Performance Measures:} Several performance measures have been derived on the confusion matrix. Among them, the most  relevant measure is the F-measure \cite{InformationTheory} defined as the harmonic mean of precision and recall. However, F-measure violates the triangle inequality, and therefore, not suitable as a metric \cite{fscorecriticize1}. Furthermore, F-measure is not symmetric in the positive and negative classes. These violations prevent the use of F-measure for comparison among classifiers in a consistent manner. Moreover, \cite{fscorecriticize2} points out that, except for accuracy, all information theoretic performance measures have undefined intervals. For example, F-measure is undefined when the number of TP is $0$ even if there are detections. Since F-measure is not  directly designed for the object detection problem, the tightness of the bounding boxes is not evaluated. AP is an information theoretic measure, too. We talk about its deficiencies in Sections \ref{sec:intro} and \ref{sec:AveragePrecision}.

\textbf{Point Multi-target Tracking Performance Metrics:} Object detection is very similar to the multi-target tracking problem. In both problems, there are multiple instances to detect, and the localization, FN and FP rates are common criteria for success. Therefore, here we review the measures and metrics used in multi-target tracking. Currently, component-based performance metrics are the accepted way of evaluating point multi-target tracking filters aiming to infer the state of each target given the measurements until the detection time. The first metric to combine the localization and cardinality (including both FP and FN) errors is the Optimal Subpattern Assignment (OSPA) \cite{OSPA}.  Following OSPA, several measures and metrics have been  proposed as its variants  \cite{OSPA,OtherOSPA1,OtherOSPA2,OtherOSPA3,OtherOSPA4,OtherOSPA5,DASA}. However, similar measures and metrics are lacking in the object detection literature, though similar performance evaluation problems are observed. 

\textbf{Setting the Thresholds of the Classifiers:} The research on the optimization of a precision-recall balanced performance measure is mostly concentrated on the F-measure. \cite{f1optimizer} considers maximizing F-measure at the inference time using plug-in rules, while \cite{SVMtrainingf1,CRFtrainingf1} offer maximization during training for support vector machines and conditional random fields. Similarly, \cite{maximizeF1Lipton} aims to find optimal thresholds for a probabilistic classifier based on maximizing the F-measure. Finally, \cite{MaximizeF1NIPS} presents a theoretical analysis of optimization of the F-measure, which also confirms the threshold-F-measure relationship depicted in \cite{maximizeF1Lipton,multilabel}.

\textbf{In summary}, we see that existing methods mostly focus on the F-measure for optimizing the thresholds for classifiers, which, however, has the aforementioned drawbacks. Moreover, F-measure is shown to be concave with respect to its inputs, number of TPs and FPs \cite{maximizeF1Lipton}, which makes the analytical optimization impossible. In addition, none of these studies have considered the object detection problem in particular, thus no localization error is directly involved for these measures. Therefore, different from the previous work, we specifically are interested in performance evaluation and optimal thresholding of the object detectors of deep learning frameworks. Moreover, we directly optimize a well-behaving function which has a smaller domain in practice in order to identify the class-specific thresholds.

\section{Average Precision: an analysis and its deficiencies}
\label{sec:AveragePrecision}
Due to space constraints, we omit the definition of AP and refer the reader to the accompanying supplementary material or \cite{PASCAL} for a definition. There exist minor differences in AP’s practical usage. For example, AP is computed by simply integrating over 11 points (that divide the entire recall domain in equal pieces) in the PASCAL VOC 2007 challenge \cite{PASCAL} whereas in MSCOCO \cite{COCO}, 101 points are used. Precision values at intermediate points are simply interpolated to prevent wiggles in the curve, by setting it to the maximum precision obtained in the interval of higher recall than the current point. While a single intersection-over-union (IoU) threshold, which is $0.5$, is used in PASCAL VOC \cite{PASCAL}; a range of IoU thresholds (from $0.5$ to $0.95$) are used in MSCOCO; the average AP over this range of IoU thresholds is also called mAP.

AP aims to evaluate the precision of the detector over the entire recall domain. Thus, it favors the methods that have precision over the entire recall domain, instead of the detectors whose RP curves are nearer to the top-right corner. In other words, AP does not compare the maximum but the overall capability/performance of the detectors.

The most important two deficiencies of AP are discussed in Section \ref{sec:intro}. In the following, we list other, more minor deficiencies.

\textbf{AP is not confidence-score sensitive}. Since the sorted list of the detections is required to calculate AP, a detector generating results in a limited interval will lead to the same AP. To illustrate, there are only $2$ detections with same confidence score in Fig. \ref{fig:differentdetectors} out of $4$ ground truths. Note that setting the confidence scores to any value(i.e. $0.01$) leads to the same AP as long as the order is preserved.

\textbf{AP does not suggest a confidence score threshold for the best setting} of the object detector. However, in a practical application, the detections are usually required to be filtered owing to response time limitations. For example, the state-of-the-art online object detector \cite{AssociationLSTM} applies a confidence score threshold of $0.8$ on the SSD method \cite{SSD} and obtains $12fps$ in this fashion.

\textbf{AP uses interpolation between neighboring recall values}, which is especially problematic for classes with very small size. For example, ``toaster” class of \cite{COCO} has 9 instances in the validation 2017 set. 
\section{Localization-Recall-Precision (LRP) Error}
\label{section:LRP}
Let $X$ be the set of ground truth boxes and $Y$ be the set of  boxes returned by an object detector. To compute $\mathrm{LRP}(X,Y_s)$, the LRP error of $Y_s$ against $X$ at a given score threshold $s$ ($0 \leq s \leq 1$) and IoU threshold $\tau$ ($0 \leq \tau<1$); first, we create a new set, $Y_s$, containing only the detections that pass the threshold $s$ and then, assign the detections in $Y_s$ to ground-truth boxes in $X$, as performed for $AP$. Once making the assignments, we compute three values: (i) $N_{TP}$, the number of true positives; (ii) $N_{FP}$, the number of false positives; (iii) $N_{FN}$, the number of false negatives. Using these quantities, the LRP error, $\mathrm{LRP}(X,Y_s)$, is defined as follows:
\small
\begin{align}
\label{eq:LRPdef}
\mathrm{LRP}(X,Y_s) := \frac{1}{Z} \left( w_{IoU} \mathrm{LRP}_{IoU}(X,Y_s)+ w_{FP} \mathrm{LRP}_{FP}(X,Y_s) + w_{FN} \mathrm{LRP}_{FN}(X,Y_s) \right),
\end{align}
\normalsize
where $Z=N_{TP}+N_{FP} +N_{FN}$ is the normalization constant; and the weights $w_{IoU}=\frac{N_{TP}}{1-\tau}$, $w_{FP}=|Y_s|$, and $w_{FP}=|X|$ control the contributions of the terms. The weights make each component easy to interpret, provide further information about the detector and prevent the total error from being undefined whenever the denominator of a single component is $0$. $\mathrm{LRP}_{IoU}$ represents the IoU tightness of valid detections as follows:
\begin{align}
\label{eq:Loc}
\mathrm{LRP}_{IoU}(X,Y_s):=\frac{1}{N_{TP}}\sum \limits_{i=1}^{N_{TP}} (1-IoU(x_i, y_{x_i})),
\end{align}
which measures the mean bounding box localization error resulting from correct detections. Another interpretation is that $1-\mathrm{LRP}_{IoU}(X,Y_s)$ is the average IoU of the valid detections.

The second component, $\mathrm{LRP}_{FP}$, in Eq. \ref{eq:LRPdef} measures the false-positives:
\begin{align}
\label{eq:Type1}
\mathrm{LRP}_{FP}(X,Y_s):= 1-Precision=1- \frac{N_{TP}}{|Y_s|}=\frac{N_{FP}}{|Y_s|},
\end{align}
and false negatives are measured by $\mathrm{LRP}_{FN}$:
\begin{align}
\label{eq:Type2}
\mathrm{LRP}_{FN}(X,Y_s):= 1-Recall=1- \frac{N_{TP}}{|X|}=\frac{N_{FN}}{|X|}.
\end{align}
FP \& FN  components together represent precision-recall of the corresponding $Y_s$ by $1-\mathrm{LRP}_{FP}(X,Y_s)$ and $1-\mathrm{LRP}_{FN}(X,Y_s)$ respectively.
Denoting the IoU between $x_i \in X$ and its assigned detection $y_{x_i} \in Y_s$ by $IoU(x_i, y_{x_i})$, the LRP error can be equally defined in a more compact form as:
\begin{align}
\label{eq:LRPdefcompact}
\mathrm{LRP}(X,Y_s):= \left( \sum \limits_{i=1}^{N_{TP}}  \frac{1-IoU(x_i, y_{x_i})}{1-\tau}+N_{FP} +N_{FN} \right) / ({N_{TP}+N_{FP} +N_{FN}}).
\end{align}
$\mathrm{LRP}$ penalizes each TP by its erroneous localization normalized by $1-\tau$ to the [0,1] interval, each FP and FN by $1$ that is the penalty upper bound. This sum of error is averaged by the total number of its contributors, i.e.,  $N_{TP}+N_{FP} +N_{FN}$. So, with this normalization, $\mathrm{LRP}$ yields a value representing the average error per bounding box in the [0,1] interval.

Overall, the ranges of total error and the components are $[0,1]$ and lower value implies better performance. At the extreme cases; $0$ for $\mathrm{LRP}$ means that each ground truth item is detected with perfect localization, and if $\mathrm{LRP}$ is $1$, then no valid detection matches the ground truth (i.e., $|Y_s|=N_{FP}$). $\mathrm{LRP}$ is undefined only when the ground truth and detection sets are both  empty (i.e., $N_{TP}+N_{FP} +N_{FN}=0$), i.e., there is nothing to evaluate.

As for the parameters, $s$ is the confidence score threshold, and $\tau$ is the IoU threshold. Since the RP pair is directly identified by the FP\&FN components, each different detection set $Y_s$ corresponds to a specific point of the RP curve. For this reason, decreasing $s$ corresponds to moving along the RP curve in the positive recall direction. $\tau$ defines minimum overlap for a detection to be validated as a TP. In other words, higher $\tau$ means we require tighter BBs. Overall, both parameters are related with the RP curve: A $\tau$ value corresponds to drawing the RP curve and  an $s$ value determines a point on the RP curve to evaluate in terms of the LRP error.

In the supplementary material, we prove that LRP is a metric.
\section{Optimal LRP (oLRP) Error: The Performance Metric and Thresholder}

Optimal LRP (oLRP) is defined as the minimum achievable LRP error with $\tau=0.5$, which makes $\mathrm{oLRP}$ parameter independent:
\begin{align}
\label{eq:OptimalLRP}
\mathrm{oLRP}:= \min_s \mathrm{LRP}(X,Y_s).
\end{align}
For ablation studies and practical requirements, different $\tau$ values can be adopted. In such cases, $\mathrm{oLRP}@\tau$ can be used to denote the Optimal LRP error at $\tau$.

oLRP searches among the confidence scores to find the best balance for competing precision-recall-IoU. The RP setting of the RP curve that oLRP  has found corresponds to the top-right part of the curve, where the optimal balanced setting resides. We call a curve \textit{sharper} than another RP curve, if its peak point at the top-right part is nearer to the $(1,1)$ RP pair. To illustrate, the RP curves in Fig. \ref{fig:differentdetectors}(d) and \ref{fig:differentdetectors}(e) are sharper than that in Fig. \ref{fig:differentdetectors}(f).

The components of $\mathrm{oLRP}$ are coined as optimal box localization ($\mathrm{oLRP}_{IoU}$),  optimal FP ($\mathrm{oLRP}_{FP}$), and optimal FN ($\mathrm{oLRP}_{FN}$) components. In this case, $\mathrm{oLRP}_{IoU}$ describes the mean average tightness for a class, and $\mathrm{oLRP}_{FP}$ and $\mathrm{oLRP}_{FN}$ together pertain to the sharpness of the curve since the corresponding RP pair is the maximum achievable performance value of the detector for this class. One can directly pinpoint the sharpness point by $1-\mathrm{oLRP}_{FP}$ and $1-\mathrm{oLRP}_{FN}$. Overall, different from AP, oLRP aims to find out the best class specific setting of the detector and it favors sharper ones that also represent better BB tightness.

Denoting oLRP error of class $c \in C$ by $\mathrm{oLRP}_c$, Mean Optimal LRP (moLRP) is defined as follows:
\begin{align}
\label{eq:MeanOptimalLRP}
{\mathrm{moLRP}}:= \frac{1}{|C|}  \sum \limits_{c\in{C}} \mathrm{oLRP}_c .
\end{align}
As in mAP, ${\mathrm{moLRP}}$ is the performance metric for the entire detector. Mean optimal box localization, FP and FN components, denoted by $\mathrm{moLRP}_{IoU}$, $\mathrm{moLRP}_{FP}$, $\mathrm{moLRP}_{FN}$ respectively, are similarly defined as the mean of the class specific components. Different from the components in oLRP, the mean optimal FP and FN components are not necessarily on the average of the RP curves of all classes due to averaging $\mathrm{moLRP}_{FP}$ (i.e., precision) with different $\mathrm{moLRP}_{FN}$ (i.e., recall) values but still provides information on the sharpness of the RP curves as shown in the experiments.

Owing to its filtering capability, $\mathrm{oLRP}$ can be used for thresholding purposes. If a problem needs an image object detector as the backbone and processing is to be completed within limited time, then only a small subset of the detections should be selected. For such methods, using an overall confidence score for the object detector is a common approach \cite{AssociationLSTM}. For such a task, oLRP identifies the class-specific best confidence score thresholds. One possible drawback of this method is that validated detections can still be too large to be processed in the desired limited time. However, by accepting larger LRP errors, higher confidence scores can be set, but again in a class-specific manner. Second practical usage of oLRP is about the deployment of the devised object detector into a platform in which confidence scores are to be discarded for user-friendliness. In such a case, one needs to set the $\tau$ threshold considering the application requirements while optimizing for the best confidence score.

In essence, calculating $\mathrm{oLRP}$ is an optimization problem. However, thanks to the smaller search space, we propose to discretize the $s$ domain into $0.01$ spaced intervals and search exhaustively in this limited space.
\section{Experimental Evaluation}
\label{section:experiments}

In this section, we analyze the parameters of LRP, represent its discriminative power on common object detectors and finally show that the class specific thresholds increase the performance of a simple online video object detector.

\textbf{Evaluated Object Detectors:} We evaluate commonly used deep object detectors; namely, Faster R-CNN, RetinaNet, and SSD. For Faster R-CNN and RetinaNet variants, we use the models by \cite{Detectron2018} and for SSD variants, the models of \cite{SSDKeras} are utilized. For the variants, we use R50, R101 and X101 while referring to the ResNet-50, ResNet-101 and RexNeXt-101 backbones respectively and FPN for feature pyramid network. All models are tested on ``MS COCO validation 2017” including 80 classes and 5000 images.

\begin{figure}[hbt!]
\centering
\includegraphics[width=1.0\textwidth]{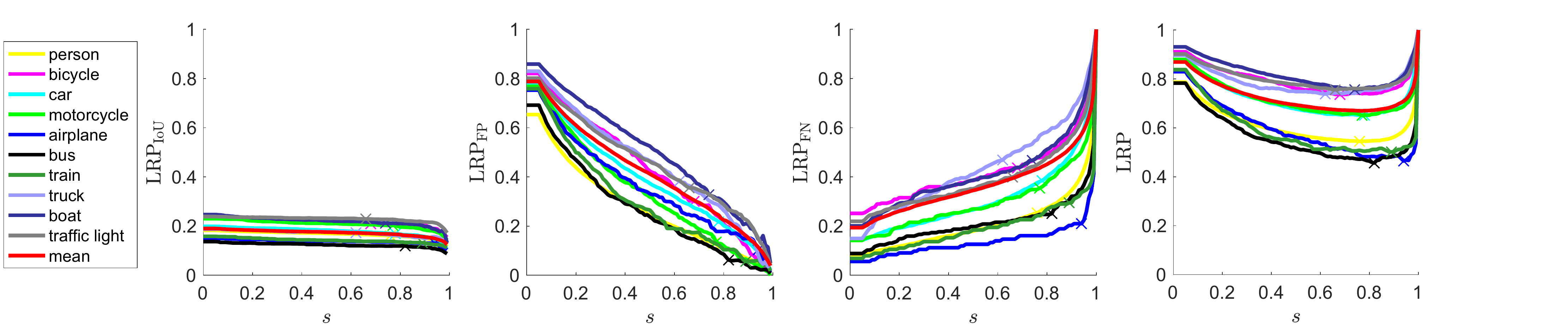}
\caption{For each class, LRP components \& total error of Faster R-CNN (X101+FPN) are plotted against $s$. The optimal confidence scores are marked with crosses. [Best viewed in color]}
\label{fig:s}
\vspace*{-0.5cm}
\end{figure}
\subsection{Analyzing Parameters $s$ and $\tau$} 
\begin{figure}
\centering
\includegraphics[width=1.0\textwidth]{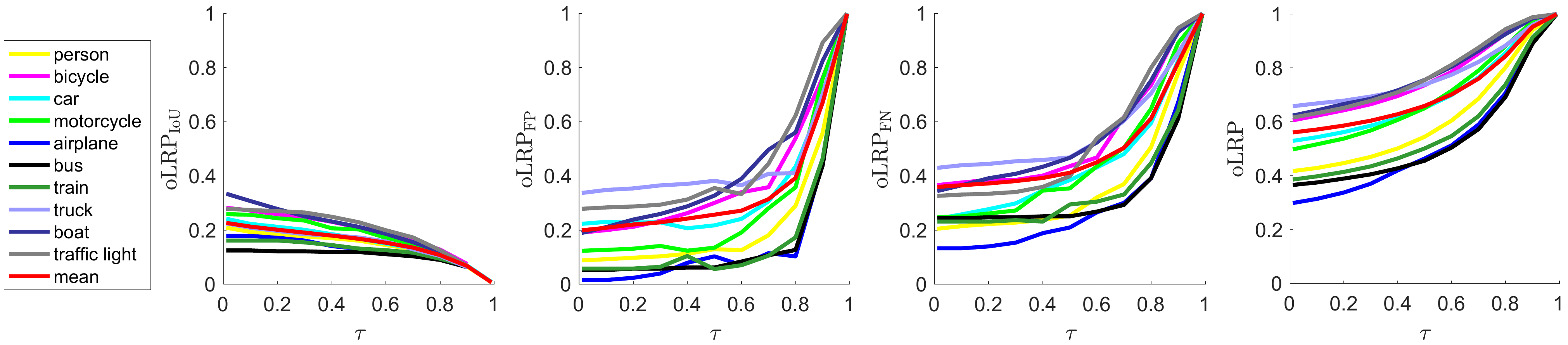}
\caption{For each class, oLRP and its components for Faster R-CNN (X101+FPN) are plotted against $\tau$. The mean represents the mean of 80 classes. [Best viewed in color]}
\label{fig:tau}
\end{figure}

Using Faster R-CNN (X101+FPN) results of the first 10 classes and mean-error for clarity, the effects of the $s$ and $\tau$ are analyzed in Fig. \ref{fig:s} and \ref{fig:tau}. We observe that box localization component is not significantly affected by increasing $s$, except for large $s$, where the error slightly decreases since the results tend to be more ``confident''. FP and FN components act in contrast to precision and recall respectively, as expected. Therefore, lower curves imply better performance for these components. Finally, the total error (oLRP) has a second-order shape. Since the localization error is not affected significantly by $s$, the behavior of the total error is mainly determined by FP and FN components, which result in the global minima of the total error to have a good precision and recall balance.

In Fig. \ref{fig:tau}, oLRP and moLRP are plotted against different $\tau$ values. As expected, larger $\tau$ values imply lower the box localization component ($\mathrm{oLRP_{IoU}}$). On the other hand, increase $\tau$ causes FP and FN components to increase rapidly, leading to higher total error (oLRP). This is intuitive since at the extreme case, i.e., when $\tau=1$, there are hardly any valid detections and all the detections are false positives, which makes oLRP to be approximately $1$. Therefore, oLRP allows measuring the performance of a detector designed for an application that requires a different $\tau$ by also providing additional information. In addition, investigating oLRP for different $\tau$ values represents a good extension for ablation studies.

\vspace{-1.1cm}

\subsection{Evaluating Common Image Object Detectors}
 \begin{table}
\caption{Performance comparison of common object detectors. R50, R101 and X101 represent the backbone networks used by ResNet-50, ResNet-101 and RexNeXt-101, respectively, and FPN refers to the feature pyramid network. $s^*_{min}$ and $s^*_{max}$ denote minimum and maximum class-specific thresholds respectively for oLRP. \label{tbl:perf_comparison}} 
 \centering
 \scriptsize 
\begin{tabular}{|l|c|c|c|c|c|c|c|c|}
\hline
&mAP&mAP@{0.5}&moLRP & $\mathrm{moLRP_{IoU}}$& $\mathrm{moLRP_{FP}}$& $\mathrm{moLRP_{FN}}$&$s^*_{min}$&$s^*_{max}$\\
\hline
SSD-300&$0.161$&$0.383$&$0.854$&$0.281$&$0.403$&$0.622$&$0.05$&$0.53$\\
SSD-512&$0.284$&$0.481$&$0.763$&$0.202$&$0.331$&$0.549$&$0.08$&$0.63$\\
Faster R-CNN (R50)&$0.348$&$0.557$&$0.714$&$0.183$&$0.292$&$0.484$&$0.18$&$0.93$\\
RetinaNet (R50+FPN)&$0.357$&$0.547$&$0.711$&$0.169$&$0.293$&$0.503$&$0.26$&$0.60$\\
Faster R-CNN (R50+FPN)&$0.379$&$0.593$&$0.689$&$0.175$&$0.259$&$0.454$&$0.41$&$0.94$\\
RetinaNet (X101+FPN)&$0.398$&$0.595$&$0.677$&$0.161$&$0.255$&$0.462$&$0.28$&$0.70$\\
Faster R-CNN (R101+FPN)&$0.398$&$0.613$&$0.673$&$0.168$&$0.255$&$0.436$&$0.37$&$0.94$\\
Faster R-CNN (X101+FPN)&$0.413$&$0.637$&$0.663$&$0.171$&$0.256$&$0.413$&$0.39$&$0.94$\\
  \hline
\end{tabular}

\vspace*{-0.5cm}
 \end{table}

\begin{wrapfigure}{r}{0.5\textwidth}
  \begin{center}
    \vspace*{-0.7cm}
    \includegraphics[width=5cm]{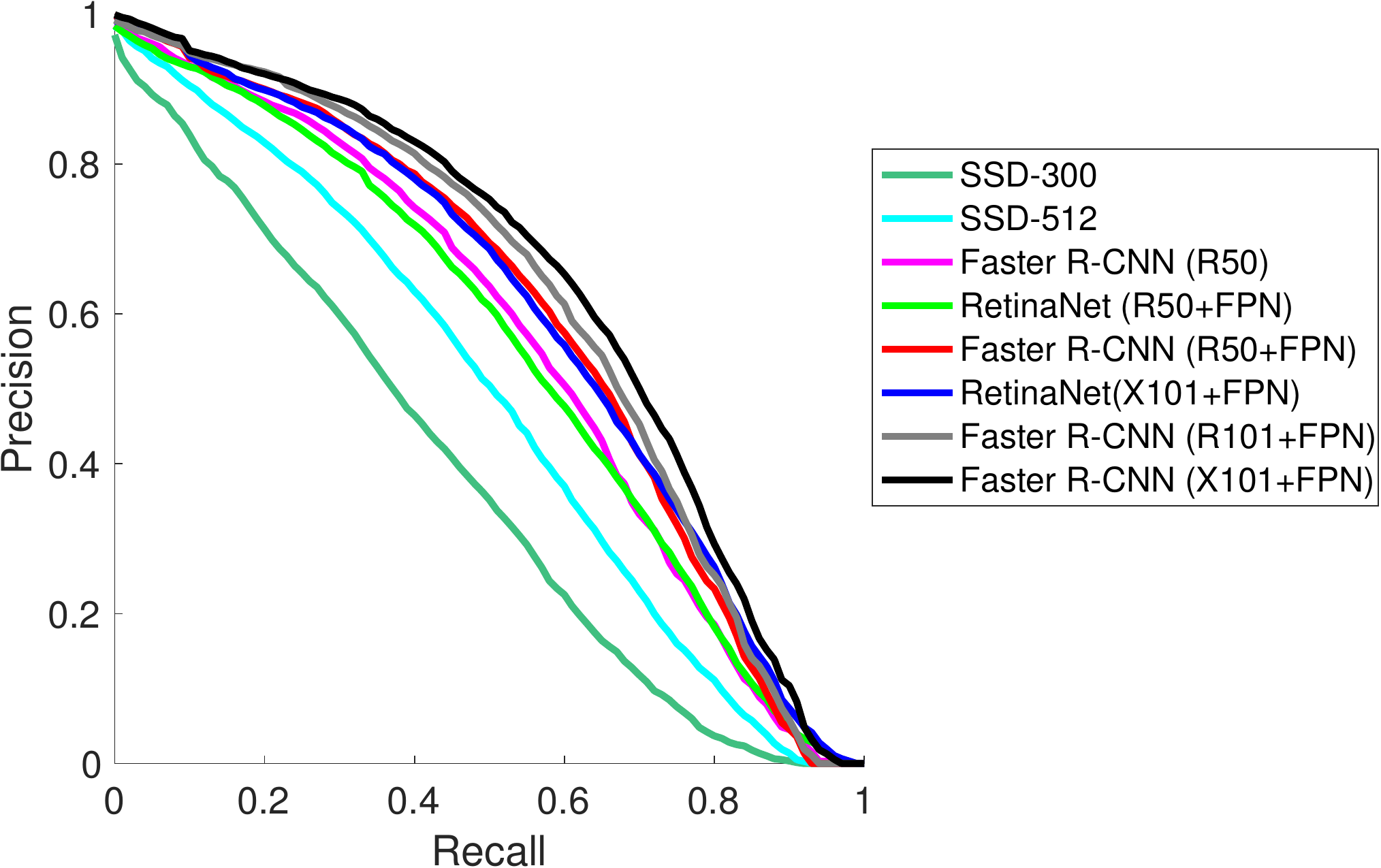}
  \end{center}
\caption{Average RP curves of the common detectors. [Best viewed in color]}
\vspace{-0.5cm}
\label{fig:AvgPRCurves}
\end{wrapfigure}

\textbf{General Overview:}  Table \ref{tbl:perf_comparison} compares the detectors using mAP as the COCO’s standard metric, mAP@0.50, moLRP and the class-specific threshold ranges. We observe that moLRP values are indicative of the known performances of the detectors. For any type of the detector, each new property (i.e., including FPN, increasing depth, using ResNext for Faster R-CNN and RetinaNet, increasing input size to 512 for SSD) decreases moLRP as expected. Moreover, the overall order is consistent with mAP except for RetinaNet (X101+FPN) and Faster R-CNN (R101+FPN), which are equal in terms of mAP; however, Faster R-CNN (R101+FPN) surpasses RetinaNet (X101+FPN) in terms of moLRP, which is discussed below. Note that  $\mathrm{moLRP_{FP}}$ and $\mathrm{moLRP_{FN}}$ values in Table  \ref{tbl:perf_comparison} are also consistent with the sharpness of the RP curves of the methods as presented in Fig. \ref{fig:AvgPRCurves}. To illustrate, Faster R-CNN (X101+FPN) has the best $\mathrm{moLRP_{FP}}$, $\mathrm{moLRP_{FN}}$ combination, corresponding to the sharpest RP curve. Another interesting example pertains to the RetinaNet (X101+FPN) and Faster R-CNN (R50+FPN) curves. For these methods, $\mathrm{moLRP_{FP}}$ and $\mathrm{moLRP_{FN}}$ comparison slightly favors Faster R-CNN (R50+FPN), which is justified by their PR curves in Fig. \ref{fig:AvgPRCurves}.

\begin{figure}
\centering
\includegraphics[width=1.0\textwidth]{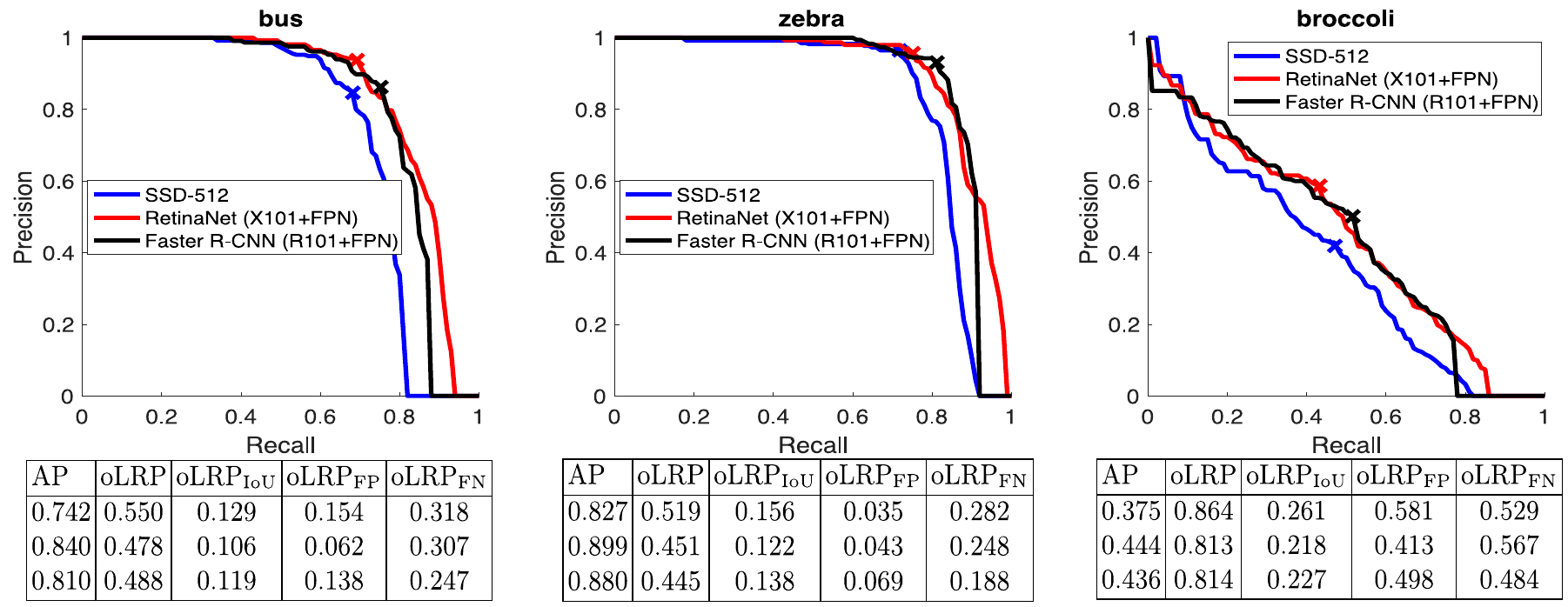}
\caption{Example RP curves representing the optimal configurations marked with crosses. The curves are drawn for $\tau=0.5$. The tables in the figures represent the performance of the methods with respect to AP and moLRP. The rows of the table correspond to SSD-512, RetinaNet (X101+FPN) and Faster R-CNN (R101+FPN) respectively.}
\label{fig:CommonPRCurves}
\end{figure}

\textbf{Class-Based Comparison and Interpreting the Components:} Now we analyze oLRP on a class-basis and look at the individual components to get a better feeling about the characteristics of methods -- see Fig. \ref{fig:CommonPRCurves}. For all three classes, oLRP is determined at the RP pairs where there exists a sharp precision decrease on the top right part of the curve. Moreover, intuitively, these pairs provide a good balance between precision and recall. Considering the FP and FN components, one can infer the structure of the curve. For all methods, the ``zebra’’ class has the sharpest RP curves which correspond to lower FP \& FN error values. For example, Faster R-CNN has $0.069$ and $0.188$ FP and FN error values, respectively. Thus, without looking at the curve, one may consider that the peak of the curve resides at $1-0.069=0.931$ precision and  $1-0.188=0.812$ recall. For the ``broccoli” curve, a less sharp one, the optimal point is at $1-0.498=0.502$ and $1-0.484=0.516$ as precision and recall respectively. Similar to ``zebra’’, these values suggest that the peak of the curve is around the center of the RP range. The localization component ($\mathrm{oLRP_{IoU}}$) shows that the tightness of the boxes for the ``bus” class is better than that of ``zebra” for all detectors even though ``zebra” has a sharper RP curve.  For RetinaNet, average IoU is $1-0.106=0.894$ and $1-0.122=0.878$ for the ``bus” and ``zebra” classes respectively. With this analysis, we also see that it is easy to compare the tightness of the boxes among the methods and classes.

\textbf{Same mAP but different behaviors, Faster R-CNN vs. RetinaNet:} Now we compare two detectors with equal AP in order to identify their characteristics using the components of  moLRP; namely, RetinaNet (X101+FPN), a single shot detector and Faster R-CNN (R101+FPN), a two-step detector. Firstly, we use the box localization component ($\mathrm{moLRP_{IoU}}$) in Table \ref{tbl:perf_comparison} to discriminate between these two detectors. The standard metric used in MS COCO aims to include the localization error by averaging over 10 mAP values. Since $1.8\%$ difference for these two detectors is present in the mAP@0.5, one can infer that RetinaNet seems to produce more tight boxes. However, this inference is possible only by examining all 10 mAP results one by one and still it is not possible to quantize this tightness. In contrast,  $\mathrm{moLRP_{IoU}}$ directly suggests that, among all the detectors in Table \ref{tbl:perf_comparison}, RetinaNet (X101+FPN) produces the tightest bounding boxes with an average tightness of $1-0.161=0.839$ in IoU terms.

Secondly, we compare the sharpness of the same two detectors, which are evidently different (Fig. \ref{fig:AvgPRCurves}). RetinaNet (X101+FPN) produces $486,108$ bounding boxes for $36,781$ annotations, whereas Faster R-CNN (R101+FPN) yields only $127,039$ thanks to its RPN method. For RetinaNet, confidence scores of $57\%$ of the detections are under $0.1$, and $87\%$ of them under $0.25$ (these values are $29\%$ and $56\%$ for Faster R-CNN), which generally causes RetinaNet to have lower or equal precision than Faster R-CNN throughout the recall domain except for the tail of the RP curve. In the tail of RetinaNet, owing to its large number of results, it has some precision even though that of Faster R-CNN drops to $0$. Fig. \ref{fig:CommonPRCurves} illustrates this phenomenon, which is best observed in the ``zebra” curve. In this curve, even though RetinaNet has higher AP than Faster R-CNN with $0.899$ to $0.880$, this AP difference originates from the large number of RetinaNet detections, which causes the better RP curve tail. This shallow curve-longer tail phenomenon is observed to be more or less valid for more than $50$ classes including the ones in Fig. \ref{fig:PRCurves}. On the other hand,  oLRP and thus moLRP do not favor these kind of detectors but the sharper ones as shown in Fig. \ref{fig:CommonPRCurves}, which causes Faster R-CNN (R101+FPN) to have lower Optimal LRP error for ``zebra’’ class.

Overall, even though RetinaNet has the best bounding box localization, Faster R-CNN (R101+FPN) with the same AP has lower mean oLRP error. Moreover, considering the RP curve of these variants, Faster R-CNN is sharper than RetinaNet as shown in Fig. \ref{fig:AvgPRCurves}. This is also validated by the components with nearly equal  $\mathrm{moLRP_{FP}}$ and difference in  $\mathrm{moLRP_{FN}}$ in favor of Faster R-CNN. Similarly, both  $\mathrm{moLRP_{FP}}$ and $\mathrm{moLRP_{FN}}$ for RetinaNet (R50+FPN) are greater than those of Faster R-CNN (R50) due to the same shallow curve-longer tail phenomenon, preventing its RP curves to be sharper. Again, what makes RetinaNet (R50+FPN) to have better performance regarding both mAP and moLRP is its strength to produce tight bounding boxes as shown in Table \ref{tbl:perf_comparison}.

\subsection{Better Threshold, Better Performance}
In this experiment, we demonstrate a use-case where oLRP helps us to set class-specific optimal thresholds such that the performance is increased compared to the naive approach of using a general threshold for all classes. To this end, we developed a simple, online video object detection framework where we use an off-the-shelf still-image object detector (RetinaNet-50 \cite{FocalLoss} trained on MS-COCO \cite{COCO}) and built three different versions of the video object detector. The first version, denoted with $B$, uses the still-image object detector to process each frame of the video independently. The second and third versions, denoted with $G$ and $S$, respectively, again use the still-image object detector to process each frame and in addition, they link bounding boxes across subsequent frames using the Hungarian matching algorithm \cite{Hungarian} and update the scores of these linked boxes using a simple Bayesian rule (details of this simple online video object detector is given in the Supplementary Material). The only difference between $G$ and $S$ is that while $G$ uses a validated threshold of $0.5$ (see $s^*$ of B in Table \ref{table2} and Fig. 1 in Supplementary Material for validation) as the confidence score threshold for all classes, $S$ uses the optimal threshold per class which achieves the oLRP error. We test these three detectors on 346 videos of ImageNet VID validation set \cite{ILSVRC} for 15 object classes which also happen to be included in MS COCO.

\begin{figure}[hbt!]
\centering
\vspace*{-0.5cm}
\includegraphics[width=0.7\textwidth]{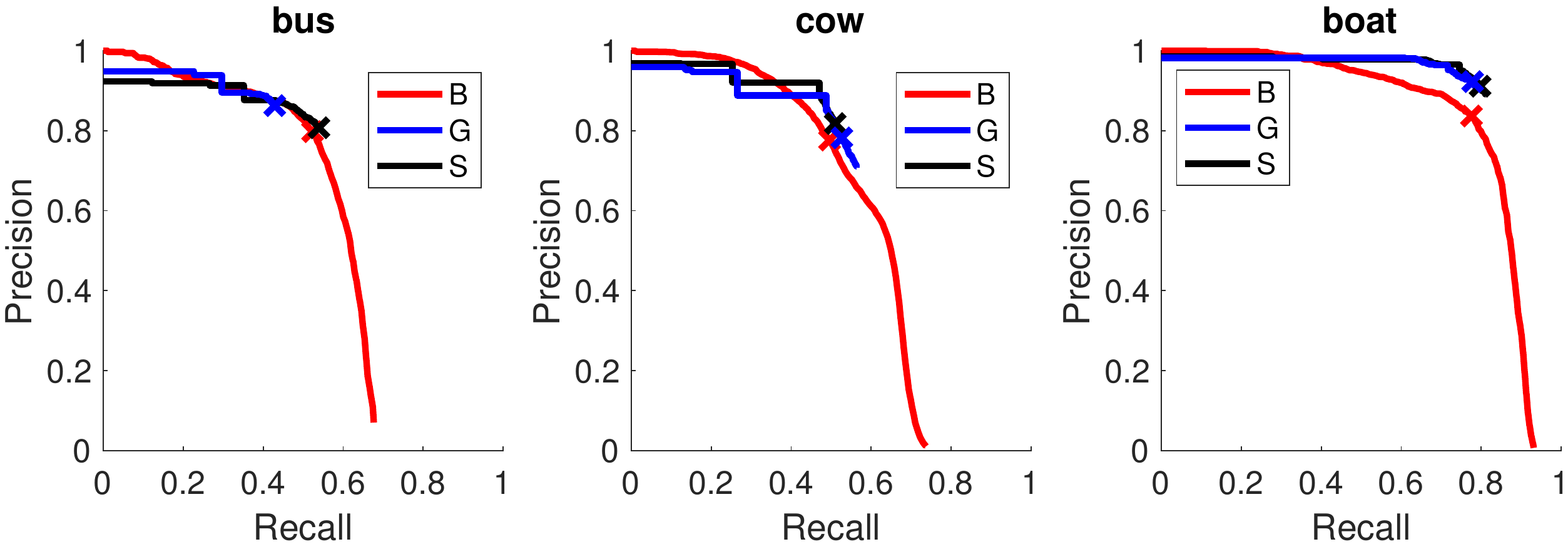}
\caption{Example RP curves of the methods. Optimal RP pairs are marked with crosses. [Best viewed in color]}
\label{fig:PRCurves}
\end{figure}

\textbf{AP vs. oLRP:} We compare $G$ with $B$ in order to represent the evaluation perspectives of AP and oLRP -- see Fig. \ref{fig:PRCurves} and Table \ref{table2}. Since $B$ is a conventional object detector, with conventional RP curves as illustrated in Fig. \ref{fig:PRCurves}. On the other hand, in order to be faster, $G$ ignores some of the detections causing its maximum recall to be less than that of $B$. Thus, these shorter ranges in the recall set a big problem in the AP evaluation. Quantitatively, $B$ surpasses $G$ by $7.5\%$ AP. On the other hand, despite limited recall coverage, $G$ obtains higher precision than $B$ especially through the end of its RP curve. To illustrate, for the ``boat” class in Fig. \ref{fig:PRCurves}, $G$ has significantly better precision after approximately between $0.5$ and $0.9$ recall even though its AP is lower by $6\%$. Since oLRP compares methods concerning their best configurations (i.e. the peak of their RP curves), this difference is clearly addressed comparing their oLRP error in which $G$ surpasses $S$ by $4.1\%$. Furthermore, the superiority of $G$ is shown to be its higher precision since FN components of $G$ and $S$ are very close while FP component of $G$ is $8.6\%$ better, which is also the exact difference of precisions in their peaks of RP curves.

Therefore, while $G$ seems to have very low performance in terms of AP, for 12 classes $G$ reaches better peaks than $B$ as illustrated by the oLRP values in Table \ref{table2}. This suggests that oLRP is better than AP in capturing the performance details of the methods.
\renewcommand{\arraystretch}{2}
\begin{table}[hbt!]
\vspace*{-0.7cm}
\caption{Comparison among $B$, $G$, $S$ with respect to AP \& oLRP and their best class-specific configurations. The mean of class thresholds are assigned as N/A since the thresholds are set class-specific and the mean is not used. } 
 \centering
 \tiny 
 \begin{tabular}{|c|l|r|r|r|r|r|r|r|r|r|r|r|r|r|r|r|r|}
 \hline
 & \multicolumn{1}{c|}{\rotatebox[origin=c]{90}{Method}} & \multicolumn{1}{c|}{\rotatebox[origin=c]{90}{airplane}}  & \multicolumn{1}{c|}{\rotatebox[origin=c]{90}{bicycle}}  & \multicolumn{1}{c|}{\rotatebox[origin=c]{90}{bird}}  & \multicolumn{1}{c|}{\rotatebox[origin=c]{90}{bus}} & \multicolumn{1}{c|}{\rotatebox[origin=c]{90}{car}}  & \multicolumn{1}{c|}{\rotatebox[origin=c]{90}{cow}}  & \multicolumn{1}{c|}{\rotatebox[origin=c]{90}{dog}}  & \multicolumn{1}{c|}{\rotatebox[origin=c]{90}{cat}} & \multicolumn{1}{c|}{\rotatebox[origin=c]{90}{elephant}}  & \multicolumn{1}{c|}{\rotatebox[origin=c]{90}{horse}}  & \multicolumn{1}{c|}{\rotatebox[origin=c]{90}{motorcycle}} & \multicolumn{1}{c|}{\rotatebox[origin=c]{90}{sheep}} & \multicolumn{1}{c|}{\rotatebox[origin=c]{90}{train}} & \multicolumn{1}{c|}{\rotatebox[origin=c]{90}{boat}} & \multicolumn{1}{c|}{\rotatebox[origin=c]{90}{zebra}} & \multicolumn{1}{c|}{\rotatebox[origin=c]{90}{mean}}\\
 \hline
 \parbox[t]{2mm}{\multirow{3}{*}{\rotatebox[origin=c]{90}{AP}}}
 &B&$\mathbf{0.681}$&$\mathbf{0.630}$&$\mathbf{0.547}$&$\mathbf{0.565}$&$\mathbf{0.555}$&$\mathbf{0.587}$&$\mathbf{0.463}$&$\mathbf{0.601}$&$\mathbf{0.661}$&$\mathbf{0.473}$&$\mathbf{0.602}$&$\mathbf{0.561}$&$\mathbf{0.713}$&$\mathbf{0.829}$&$\mathbf{0.816}$&$\mathbf{0.619}$\\
 &G&$0.621$&$0.445$&$0.492$&$0.398$&$0.417$&$0.510$&$0.416$&$0.568$&$0.588$&$0.441$&$0.571$&$0.547$&$0.600$&$0.769$&$0.765$&$0.544$\\
 &S&$0.645$&$0.535$&$0.500$&$0.485$&$0.419$&$0.492$&$0.434$&$0.569$&$0.589$&$0.444$&$0.573$&$0.545$&$0.609$&$0.792$&$0.782$&$0.561$\\
  \hline  
  \parbox[t]{2mm}{\multirow{3}{*}{\rotatebox[origin=c]{90}
 {oLRP}}}&B&$0.627$&$0.776$&$0.718$&$0.702$&$0.759$&$0.692$&$0.728$&$0.700$&$0.625$&$0.723$&$0.692$&$0.677$&$\mathbf{0.583}$&$0.594$&$0.436$&$0.669$\\
 &G&$0.606$&$0.783$&$0.691$&$0.727$&$\mathbf{0.758}$&$0.679$&$0.714$&$\mathbf{0.697}$&$0.614$&$\mathbf{0.699}$&$\mathbf{0.654}$&$\mathbf{0.648}$&$0.586$&$0.553$&$0.432$&$0.656$\\
 &S&$\mathbf{0.603}$&$\mathbf{0.762}$&$\mathbf{0.687}$&$\mathbf{0.688}$&$0.759$&$\mathbf{0.678}$&$\mathbf{0.712}$&$\mathbf{0.697}$&$\mathbf{0.613}$&$0.701$&$0.655$&$0.649$&$\mathbf{0.583}$&$\mathbf{0.551}$&$\mathbf{0.425}$&$\mathbf{0.651}$\\
 \hline
 \parbox[t]{2mm}{\multirow{3}{*}{\rotatebox[origin=c]{90}{$\mathrm{oLRP_{IoU}}$}}}&B&$0.182$&$0.271$&$0.169$&$0.177$&$0.207$&$0.145$&$0.166$&$0.203$&$0.170$&$0.155$&$0.192$&$0.154$&$0.159$&$0.199$&$0.128$&$0.179$\\
 &G&$0.181$&$0.258$&$0.170$&$0.160$&$0.207$&$0.151$&$0.165$&$0.200$&$0.170$&$0.160$&$0.195$&$0.155$&$0.156$&$0.195$&$0.128$&$0.177$\\
 &S&$0.186$&$0.270$&$0.170$&$0.173$&$0.207$&$0.148$&$0.170$&$0.200$&$0.170$&$0.160$&$0.194$&$0.155$&$0.159$&$0.197$&$0.131$&$0.179$\\
 \hline
  \parbox[t]{2mm}{\multirow{3}{*}{\rotatebox[origin=c]{90}{$\mathrm{oLRP_{FP}}$}}}&B&$0.080$&$0.228$&$0.300$&$0.203$&$0.303$&$0.224$&$0.242$&$0.248$&$0.095$&$0.246$&$0.158$&$0.141$&$0.099$&$0.163$&$0.034$&$0.184$\\
 &G&$0.006$&$0.116$&$0.174$&$0.137$&$0.311$&$0.218$&$0.229$&$0.279$&$0.071$&$0.221$&$0.049$&$0.078$&$0.091$&$0.077$&$0.016$&$0.142$\\
 &S&$0.087$&$0.226$&$0.184$&$0.193$&$0.320$&$0.182$&$0.269$&$0.283$&$0.075$&$0.231$&$0.084$&$0.078$&$0.110$&$0.089$&$0.030$&$0.163$\\
 \hline
  \parbox[t]{2mm}{\multirow{3}{*}{\rotatebox[origin=c]{90}{$\mathrm{oLRP_{FN}}$}}}&B&$0.383$&$0.427$&$0.478$&$0.477$&$0.499$&$0.504$&$0.533$&$0.394$&$0.395$&$0.540$&$0.448$&$0.494$&$0.344$&$0.224$&$0.220$&$0.424$\\
 &G&$0.359$&$0.523$&$0.480$&$0.571$&$0.493$&$0.473$&$0.512$&$0.372$&$0.388$&$0.494$&$0.415$&$0.467$&$0.360$&$0.221$&$0.227$&$0.424$\\
 &S&$0.326$&$0.389$&$0.489$&$0.461$&$0.488$&$0.490$&$0.480$&$0.369$&$0.385$&$0.493$&$0.406$&$0.468$&$0.339$&$0.203$&$0.202$&$0.398$\\
 \hline
   \parbox[t]{2mm}{\multirow{3}{*}{\rotatebox[origin=c]{90}{$s^{*}$}}}&B&$0.38$&$0.31$&$0.44$&$0.27$&$0.49$&$0.61$&$0.42$&$0.49$&$0.49$&$0.52$&$0.45$&$0.51$&$0.41$&$0.45$&$0.31$&N/A\\
 &G&$0$&$0.69$&$0.97$&$0.68$&$0$&$0.96$&$0.48$&$0.70$&$0.33$&$0.64$&$0.60$&$0.84$&$0.59$&$0.90$&$0$&N/A\\
 &S&$0$&$0.54$&$0.98$&$0.45$&$0$&$0.91$&$0.49$&$0.64$&$0.39$&$0.58$&$0.63$&$0.85$&$0.55$&$0.89$&$0.54$&N/A\\
 \hline 
 \end{tabular}
\label{table2} 
 \end{table}

\textbf{Effect of the Class-specific Thresholds:} Compared to $G$, owing to the class-specific thresholds, $S$ has $2.3\%$ better mAP and $0.6\%$ better moLRP as shown in Table \ref{table2}. However, since the mean is dominated by $s^*$ around $0.5$, it is better to focus on classes with low or high $s^*$ values in order to grasp the effect of the approach. The ``bus” class has the lowest $s^*$ with $0.27$. For this class, $S$ surpasses $G$ by $8.7\%$ in AP and $4.1\%$ in oLRP. This performance increase is also observed for other classes with very low thresholds, such as ``airplane”, ``bicycle” and ``zebra”. For these classes with lower thresholds, the effect of class-specific threshold on the RP curve is to stretch the curve in the recall domain (maybe by accepting some loss in precision) as shown in the ``bus” example in Fig. \ref{fig:PRCurves}. Not surprisingly, ``cow” is one of the two classes for which AP of $S$ is lower since its threshold is the highest and thereby causing recall to be more limited. On the other hand, regarding oLRP, the result is not worse since this time the RP curve is stretched through the positive precision, as shown in Fig. \ref{fig:PRCurves}, allowing better FP errors. Thus, in any case, lower or higher, the threshold setting method aims to discover the best RP curve. There are four classes in total for which $G$ is better than $S$ in terms of oLRP. However, note that the maximum difference is $0.2\%$ in oLRP and these are the classes with thresholds around $0.5$. These suggest that choosing class-specific thresholds rather than the common general thresholding approach increases the performance of the detector especially for classes with low or high class-specific thresholds.
\section{Discussion and Conclusion}
\label{section:conclusions}
We have introduced a novel performance metric, LRP, and the best LRP value as Optimal LRP that have significant practical applications, compared to AP:

\textbf{Optimal LRP captures the behaviour of the detector.}  As illustrated in Fig. \ref{fig:differentdetectors}, LRP addresses the weaknesses/strengths of the detectors by representing both the peak values of the RP curves and their localization capability.

\textbf{Since LRP filters the detections lower than $s$, confidence scores affect LRP with its $s$ parameter.} This parameter dependency is discarded in Optimal LRP by identifying the best confidence score.

\textbf{Optimal LRP compares different detectors and the different configurations of the same detector considering practical applications.} To illustrate, the total error of detectors in Fig. \ref{fig:differentdetectors}(a), (b), (c) are $0.5$, $0.5$ and $0.93$ respectively. Since the detector (c) has both of the problems of detector (a) and detector (b) also with a significant BB tightness problem, it has the highest oLRP. Even if the detector (c) had the perfect localisation as (a) and (b), total error would be $0.66$, again the worst due to both precision and recall problems. Also, the best configurations of the detectors in Fig. \ref{fig:differentdetectors} are identified by Optimal LRP and marked on each RP curve.

\textbf{$1- \mathrm{oLRP_{IoU}}$ defines the mean box localization accuracy of a detector.} For this reason, it alleviates estimating the AP@$\tau$ with different $\tau$ values.

\textbf{Optimal LRP error can even evaluate the detection result on a single image without any interpolation.}


\textbf{Supplementary Material:}  The paper is accompanied by supplementary material, containing a detailed definition of AP, a result on the distribution of confidence thresholds, a description of the online detector and the proof that LRP is a metric.

\appendix
\bibliographystyle{splncs}
\bibliography{references}
\end{document}


\pagestyle{headings}
\mainmatter
\def\ECCV18SubNumber{1134}  

\title{Supplementary Material for Localization Recall Precision (LRP): A New Performance Metric for Object Detection} 

\titlerunning{LRP: A New Performance Metric for Object Detection}

\authorrunning{Kemal Oksuz, Baris Can Cam, Emre Akbas and Sinan Kalkan}

\author{Kemal Oksuz, Baris Can Cam, Emre Akbas and Sinan Kalkan}
\institute{Department of Computer Engineering, Middle East Technical University \\ \texttt{\{kemal.oksuz,can.cam,eakbas,skalkan\}@metu.edu.tr}}

\maketitle 
\begin{abstract}

In this document, we provide supplementary material that were omitted in the original paper due to space constraints.

\keywords{Supplementary Material}
\end{abstract}
\section{Definition of AP}
AP is simply defined as the area under the recall-precision (RP) curve. For completeness, a description of how the RP curve is obtained follows. First, the object detector is run over the testing images and returns a list of bounding boxes with associated confidence scores. These scores are compared to a specific threshold $T$ and the boxes having scores larger than T are considered the object detector’s prediction about the presence and location of the objects. A predicted box P is deemed a``true positive’’ if there is a ground truth box G such that the intersection area of P and G divided by the union area of P and G is larger than a given intersection-over-union (IoU) threshold (which is typically set to $0.5$). Otherwise, P is deemed a ``false positive.’’ Next, recall and precision are computed for the specific threshold T. Recall is the hit rate, the number of true positives divided by the number of all ground truth boxes in the testing set. Precision is the number of true positives divided by the number of all predicted boxes (i.e., the sum of true positives and false positives). By systematically varying the value of $T$, one  obtains a recall-precision curve. The area under this curve is called ``average precision’’ (AP). It can take values in the range [0, 1]. To report the performance over many object classes, AP is computed per class and then they are averaged to yield the “mean average precision” or “mAP” for short.

\section{Analysis of Threshold Distributions}
\begin{figure}
\centering
\includegraphics[width=1.0\textwidth]{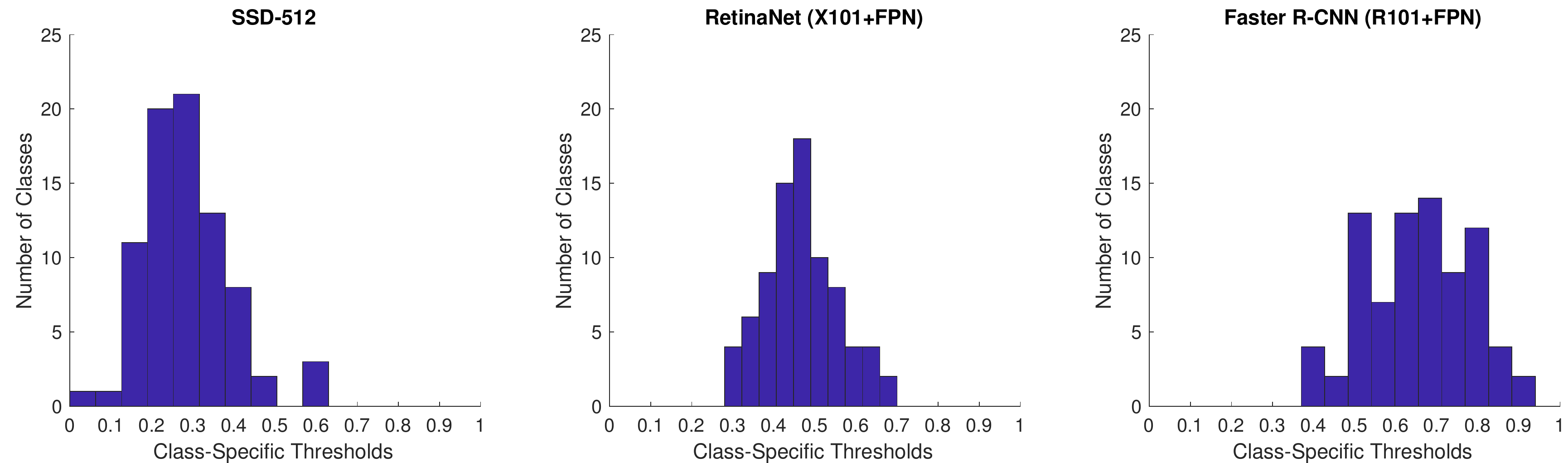}
\caption{The histograms representing the distributions of the class-specific thresholds for different methods.}
\label{fig:histogram}
\end{figure}

Considering the distributions of the class-specific thresholds in Figure \ref{fig:histogram} and $s^*_{min}$ and $s^*_{max}$, the minimum and maximum class-specific thresholds respectively in Table 1, the range of these thresholds is shown to be significantly large and varying. These observations lead to two results: Firstly, for different types of detectors, the meaning of the confidence scores are different. For example, the most confident settings of SSD-512 accumulated around $0.3$ whereas for Faster R-CNN, it is around $0.7$ as shown in Figure \ref{fig:histogram}. Secondly, setting a specific threshold for the entire detector affects each of the classes very differently. For the classes with lower optimal thresholds, the detector may very seldom produce their results, while for the classes with larger thresholds, there are many FPs. Thus, we conclude that the thresholds of the detector are to be set accordingly for each class. 

\section{A Simple Online Video Object Detector}
There are two online video object detectors: $G$ and $S$ which respectively use the general thresholding approach with $0.5$ as threshold, and the class-specific thresholds. For each of the online detectors, at each time interval, the detection BBs of previous and current frames are associated using the Hungarian algorithm \cite{Hungarian} considering a box linking function and the confidence scores of associated BBs of the current frame are updated using the score distributions of both of the BBs. Since an online tracker, specifically \cite{CFCF}, is also used in our method, we use the L1 norm of the difference of confidence score distributions of neighboring frames and the IoU overlap of the tracker prediction and the detection at current frame. While choosing this box linking score, we inspired from the tube linking score of \cite{DetectToTrack}. The updated score is estimated using the Bayes Theorem such that the prior is the updated tubelet score in the previous frame and likelihood is the currently associated high confidence detection with that tubelet. In such an update method, even though the updated scores converge to $1$ quickly, which is bad for lower recall, precision increases in larger recall portions. Also, we call a BB as ``dominant object’’ if its updated score increases by $0.2$. In order to increase the recall, the disappearance of a ``dominant object’’ is closely inspected by using the tracker again to predict the possible location, then the cropped region is classified by class-wise binary classifiers (object vs. background).

\section{Proof that LRP is a Metric}
\label{section:appendix}

In this section, we prove that LRP is a metric using a reduction from DASA \cite{DASA}, a proven metric. Firstly, Theorem \ref{theorem:IoU} proves that $1-IoU(x,y)$ is a metric, which is a requirement for metricity of LRP.

\begin{theorem}
\label{theorem:IoU}
$1-IoU(x,y)$ is a metric.
\end{theorem}
\begin{proof}
The identity and symmetry axioms are trivial. Triangle inequality is defined in Equation \ref{eq:triangle1} and reorganized in Equation \ref{eq:triangle2}, which is a fact for any arbitrary BBs $x,y,z$.
\begin{align}
\label{eq:triangle1}
1-IoU(x,y) &\leq 1-IoU(x,z)+1-IoU(z,y) \\
\label{eq:triangle2}
IoU(x,z)+IoU(z,y)-IoU(x,y) &\leq 1
\end{align}
\end{proof}
\begin{theorem}
LRP is a metric.
\end{theorem}
\begin{proof}
We show that, LRP is derived from DASA \cite{DASA}, a proven metric. Including a norm parameter $p$, a cutoff parameter $c$ DASA is defined in Equation \ref{eq:DASA1}. We denote an arbitrary metric by $d(x_i,y_{x_i})$ and $l=\max(|X|,|Y|)$.
\begin{align}
\label{eq:DASA1} 
&\bar{e}^{(c)}_p(X,Y):=\\  
&\frac{l}{Z} \left( \frac{N_{TP}}{l} \left(\frac{1}{ N_{TP}} \sum \limits_{i=1}^{|Y|} \mathbb{I}[d(x_i,y_{x_i}) \leq c]  d(x_i,y_{x_i})^p \right)+ \left(\frac{c^p  N_{FP}}{l} \right) +\left( \frac{c^p  N_{FN}}{l} \right) \right)^{1/p} \nonumber
\end{align}
Setting $p=1$, $c=1-\tau$ and $d(x_i,y_{x_i})=1-IoU(x_i,y_{x_i})$ thanks to Theorem 1;
\begin{align}
= &\frac{l}{Z} \left(  \frac{N_{TP}}{l} \left(\frac{1}{N_{TP}} \sum \limits_{i=1}^{|Y|}  \mathbb{I}[1-IoU(x_i,y_{x_i})) \geq 1-\tau]  (1-IoU(x_i,y_{x_i}))) \right) \right. \\
&\left. + \left(\frac{(1-\tau) N_{FP} }{l} \right) +\left( \frac{(1-\tau)N_{FN}}{l} \right) \right) \nonumber
\end{align}
Reorganizing the predicate in the Iverson bracket, we have $IoU(x_i,y_{x_i})) \leq \tau$, which exactly defines the validation limit of the TPs. Finally, In Equation \ref{eq:LRPderivation}, the LRP metric is derived by discarding the Iverson bracket by setting the summation upper bound to $N_{TP}$ for clarity, proposing component weights ($w_{IoU}$, $w_{FP}$, $w_{FN}$) to make them easily interpretable and normalizing the total metric by $1-\tau$ to set the upper bound to $1$. Note that, dividing by a constant does not violate metricity. Thus, since LRP is reduced from DASA, it is also a metric.
\begin{align}
\label{eq:LRPderivation}
= &\frac{1}{Z} \left( w_{IoU} \frac{1}{N_{TP}}\sum \limits_{i=1}^{N_{TP}} (1-IoU(x_i, y_{x_i}))+ w_{FP} \frac{N_{FP}}{|Y_s|} + w_{FN} \frac{N_{FN}}{|X|} \right) \\
= &\mathrm{LRP}(X,Y)
\end{align}
\end{proof}

\bibliographystyle{splncs}
\bibliography{references}